\documentclass[11pt]{article}
\usepackage[letterpaper,margin=0.85in]{geometry}
\usepackage{graphicx}
\usepackage{booktabs}
\usepackage{array}
\usepackage{xcolor}
\usepackage{hyperref}
\usepackage{microtype}
\usepackage{caption}
\usepackage{enumitem}
\usepackage{float}
\definecolor{PlatoBlue}{HTML}{1F4D78}
\hypersetup{colorlinks=true,linkcolor=PlatoBlue,urlcolor=PlatoBlue,citecolor=PlatoBlue}
\graphicspath{{figures/}}

\title{Plato-Bio: verification-first biological novelty screening with temporal rediscovery and structural benchmarks}

\author{Stefan G. Creadore\\\small Praxa Labs, an independent open-source research initiative, United States}

\date{}

\begin{document}

\maketitle

\begin{center}\small Correspondence: \href{mailto:stefan@nex-t1.ai}{stefan@nex-t1.ai} \quad ORCID: \url{https://orcid.org/0000-0003-2268-053X}\\arXiv primary category: cs.AI (Artificial Intelligence) \quad Cross-list: q-bio.QM (Quantitative Methods)\end{center}

\begin{abstract}

Large language model research agents can connect literature retrieval, analysis code, and manuscript preparation, but coherent output does not establish scientific validity. We developed Plato-Bio, a biology-routed extension of the open Plato/Denario architecture that couples explicit workflow states with provenance records, citation checks, claim-to-evidence links, scoped file writes, and publication gates. A source audit identified and repaired three defects that could distort evaluation: loss of task domain in the default factory, omission of declared method signals from scoring, and evidence sidecars that lacked the drafted-claim denominator. On the current clean revision, the full Python suite completed with 931 passes, six skips, and no failures or errors; targeted biology, genomics, evidence/citation, and adversarial-safety suites likewise completed without failure.

We evaluated two narrow use cases. In a frozen historical rediscovery task, independent pre-1986 literature bridges ranked the later-studied relation between fish oil and Raynaud phenomenon first; TF--IDF ranked it second and corpus frequency third. This single curated task measures retrospective ranking, not prospective discovery. In a separate comparison of AlphaFold models with experimental structures for 15 human proteins, 11 targets had high-confidence-core C$\alpha$ RMSD below 1 \AA{} (median 0.501 \AA{}). Four targets exceeded 2 \AA{}, and confidence masking reduced the SUMO1 discrepancy from 16.61 to 2.58 \AA{} over 74 residues. The workflow emitted 27 traceable discrepancy regions, all retained as unvalidated hypotheses. Plato-Bio therefore provides reproducible software contracts and auditable screening baselines; broader claims of agent efficacy or biological novelty require preregistered evaluation, independent review, and prospective validation.

\end{abstract}

\textbf{Keywords:} scientific agents; bioinformatics; biological novelty; literature-based discovery; temporal rediscovery; reproducibility; AlphaFold

\section{Introduction}

Automation has long been used to make parts of scientific inference more explicit and repeatable. The Robot Scientist ``Adam,'' for example, coupled hypothesis generation, experimental planning, laboratory automation, and interpretation in functional genomics ~\cite{ref1}. Recent language-model systems extend automation to literature review, code generation, experiment execution, and scientific writing. The AI Scientist ~\cite{ref2}, Agent Laboratory ~\cite{ref3}, and the Denario project ~\cite{ref4} demonstrate broad agentic workflows that can generate or assemble research artifacts. These systems are valuable prototypes, but their ability to produce polished prose creates a risk: output quality can be mistaken for evidentiary quality.

This distinction is particularly important in computational biology. Biological analyses frequently depend on versioned reference assemblies, database accessions, coordinate conventions, chain or isoform mapping, statistical assumptions, and data-use conditions. A research agent must therefore preserve not only text but also the relationship between claims, sources, calculations, and executable artifacts. FAIR data principles similarly emphasize findability, accessibility, interoperability, and reuse as properties of scientific objects rather than rhetorical descriptions ~\cite{ref5}.

Plato is a maintained fork and extension of the open Denario/Plato codebase. The prior Denario preprint describes a broad multi-agent assistant and reports expert assessment of generated papers across several disciplines ~\cite{ref4}. The present work is not a duplicate or renamed version of that manuscript. It focuses on fork-specific validation infrastructure, measurement defects discovered during a source-level audit, and a new, compact structural-biology case study whose inputs and outputs are shipped in the repository.

We make five contributions. First, we describe a biology-routed workflow in which literature sources, keyword extraction, execution defaults, and journal choices are represented by an explicit domain profile. Second, we document evidence and safety controls that create inspectable artifacts: citation-validation reports, claim/evidence matrices, scientific-consistency checks, and run manifests. Third, we repair and regression-test measurement defects that would otherwise distort evaluation. Fourth, we implement a frozen temporal rediscovery benchmark with explicit cutoffs, evidence bridges, abstention, baselines, and task-level uncertainty, aligning the evaluation design with ScienceAgentBench, BioDSA-1K, and BixBench [11--13]. Fifth, we expand the deterministic AlphaFold-to-experiment screen to 15 targets with source hashes, confidence-masked and whole-chain statistics, residue-level data, and hypothesis-only discrepancy regions. Our central claim remains narrow: the repository implements and tests these contracts and reproduces these benchmark cases. We do not infer autonomous scientific validity from them.

\section{Methods}

\subsection{Study design and claim boundary}

We conducted a source-level architecture audit, repaired measurement defects whose behavior contradicted documented contracts, ran deterministic validation suites, and executed two distinct evaluation lanes. The first is a frozen temporal rediscovery benchmark for literature-derived candidate ranking. The second is an AlphaFold-to-experiment structure screen. The original three-globin panel was retained, and a 15-target expansion was declared in \nolinkurl{preprint/experiments/diverse_structure_panel.json} before that run. Neither benchmark used an LLM or selected successful outputs after observing results.

Software tests were interpreted as contract verification only. A passing test indicates that an asserted behavior was observed in the test environment; it does not estimate biological accuracy, citation precision in open-ended use, or paper quality. Live provider tests requiring credentials or optional cloud services were reported as skipped rather than converted into successful or zero-valued scientific outcomes.

\subsection{System architecture}

Plato represents the research process as explicit LangGraph state machines rather than a single monolithic prompt. The public workflow accepts a data description, develops an idea, produces a method, executes an analysis, and drafts a manuscript. The paper graph orders section drafting before citation validation, scientific consistency checks, claim extraction, evidence linking, reviewer-role critiques, aggregation, and a bounded redraft decision. The reviewer roles use the configured drafting model in the current implementation and are therefore self-critique roles, not independent reviewers.

The biology domain profile selects PubMed, Europe PMC, OpenAlex, Crossref, DOAJ, DataCite, OpenCitations, and Semantic Scholar as retrieval sources; MeSH-based keyword extraction; PubMed as the novelty corpus; and a local Jupyter executor by default. Optional genomics adapters prepare typed, permission-gated operations for GenomeKit, ZIPPY, Paragraph, ExpansionHunter Denovo, Gauchian, and Illumina Connected Analytics. Most adapters validate requirements and prepare external commands or requests; their presence does not imply that the external software or licensed data are locally installed.

\begin{figure}[H]
\centering
\includegraphics[width=0.96\textwidth]{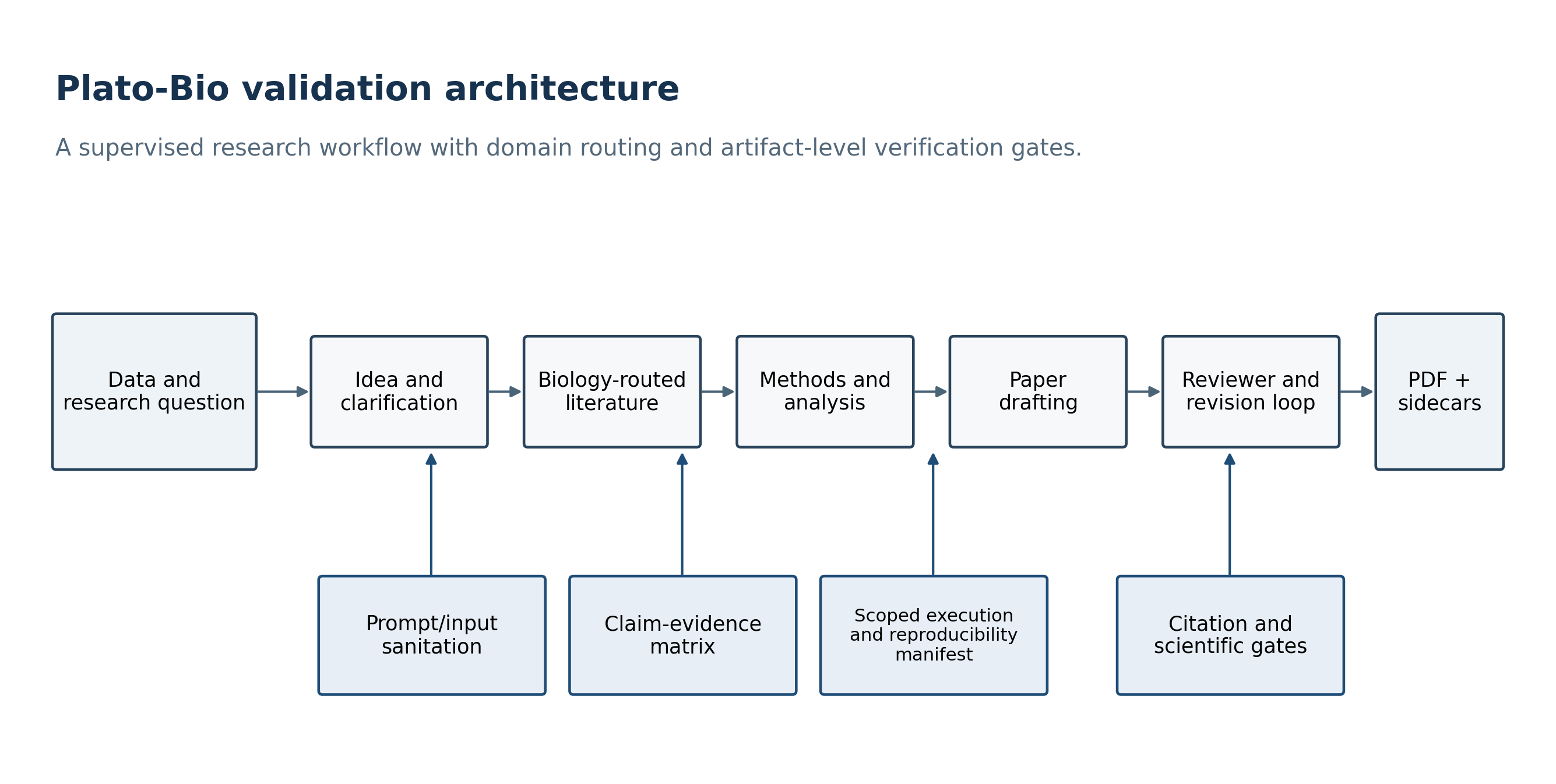}
\caption{Plato-Bio architecture and verification boundaries.}
\label{fig:figure-1}
\end{figure}

\subsection{Evidence, provenance, and publication controls}

Retrieved material is screened and wrapped as external content before entering prompts. The citation node resolves structured references and writes a validation report. A configured publication gate requires references to be present and applies a validation threshold; the threshold is a policy setting, not an observed accuracy estimate. The scientific verifier checks for artifact presence, unsupported quantitative language, provenance language, and validation status. These checks are heuristic consistency controls, not independent re-execution of every analysis.

The claim/evidence stage distinguishes drafted claims from source-derived claims and classifies source relationships as supporting, refuting, neutral, or unclear. The corrected JSONL sidecar stores the drafted Claim rows and EvidenceLink rows in the same documented stream, preserving both the denominator and support links needed to calculate unsupported-claim rate. Run-manifest primitives record workflow, timestamps, domain, repository and project hashes, models, prompts, seeds, sources, token counts, and cost when those values are supplied by the calling path.

\subsection{Measurement-validity repairs}

Three defects were treated as study-blocking because each could change an evaluation result without changing system behavior.

\begin{enumerate}[leftmargin=*]

\item The default evaluation factory constructed every task with Plato's default astronomy profile. We changed the factory to pass \nolinkurl{task.domain}, so the shipped protein task uses the biology profile.

\item Golden tasks declared \nolinkurl{expected_method_signals}, but the evaluator did not score them. We added \nolinkurl{method_signal_recall}, calculated only from the method artifact, and included it in aggregate summaries.

\item The evidence producer wrote only links while the evaluator and dashboard expected mixed claim/link rows. When no Claim rows were present, unsupported-claim rate defaulted to 0.0. We changed the producer to persist drafted claims before links, including when no source claims are available.

\end{enumerate}

Each repair was covered by a failing regression test before the implementation change and then verified with the relevant targeted suites.

\subsection{Deterministic software validation}

The repository-local script \nolinkurl{preprint/experiments/run_software_validation.py} ran five suites with the checked-in Python environment: biology-domain behavior, genomics adapters, evidence/citation controls, adversarial safety, and the complete Python test tree. It invoked pytest with JUnit XML output, parsed tests/failures/errors/skips, and wrote a JSON report containing the command, commit, platform, Python version, and timing. The targeted suites overlap with the full suite; their counts are presented to show control-specific coverage and must not be summed.

The complete suite includes unit, trajectory, safety, and opt-in integration tests. Skips correspond to missing optional SDKs or credentials for E2B, Modal, PostgreSQL, Hugging Face, and one platform-specific path case. No skipped integration was reported as validated.

\subsection{Benchmark alignment and external task catalog}

We used current scientific-agent benchmarks as design constraints rather than treating generic text similarity as an efficacy measure. ScienceAgentBench uses 102 executable tasks from 44 peer-reviewed publications and expert validation; BioDSA-1K separates hypothesis decisions, evidence alignment, reasoning, executability, and non-verifiable cases; and BixBench emphasizes long, multi-step biological analysis with open-answer interpretation [11--13]. Plato-Bio therefore reports executable or machine-readable artifacts, explicit baselines and ablations, abstention, provenance, task-level metrics, and failure boundaries. We also pinned and hash-verified the public CompBioBench v1 task catalog. Its 100 tasks span epigenomics, genomics, machine learning, population genetics, single-cell, spatial, structure, and transcriptomics. This release imports task metadata only; it does not claim CompBioBench performance because the corresponding datasets and agent runs were not executed.

\subsection{Temporal rediscovery and evidence-bridge ranking}

Each temporal task freezes an exact cutoff date, anchor concept A, candidate concepts C, a later validation publication, and pre-cutoff records with publication dates, identifiers, URLs, and manually declared concepts. Records dated on or after the cutoff cause a hard failure; missing dates also fail. Duplicate DOI/PMID records are collapsed, and records with prompt-injection signals are quarantined. A candidate can be labeled a temporally novel candidate only when no pre-cutoff record directly joins A and C and at least two different records form an A--B and B--C path. Direct A--C records are labeled known pre-cutoff; candidates without a bridge abstain as unsupported. These labels describe corpus state and do not assert biological truth.

Candidates were ranked under four declared conditions: corpus frequency, TF--IDF relevance, unweighted A--B/B--C bridge support, and an evidence-aware score. The evidence-aware condition combines normalized independent bridge pairs (0.45), TF--IDF relevance (0.25), source diversity (0.20), and provenance completeness (0.10), with a penalty of 1.0 for direct prior art. Primary endpoints were mean reciprocal rank, Recall@1, Recall@10, and false-novelty rate on known controls. Ninety-five percent intervals were computed with 10,000 task-level bootstrap resamples. Intervals from the one-task historical pilot are degenerate and are not population uncertainty estimates.

The engineering smoke contains five explicitly synthetic tasks and tests only measurement behavior. The historical pilot freezes six pre-1986 PubMed records plus a declared negative-control record. The bridge joins Raynaud phenomenon to blood viscosity through a 1976 report and fish oil to lower blood viscosity through a 1985 report ~\cite{ref14,ref15}. The held-out validation is a 1989 double-blind controlled study of fish-oil supplementation in Raynaud phenomenon ~\cite{ref16}. Nifedipine and prostaglandin E1 were included as known pre-cutoff controls because direct Raynaud-treatment records were already present. Record text in the fixture is concise curator-written paraphrase; PubMed identifiers and source URLs preserve traceability.

\subsection{Structural-biology case study}

Experimental coordinates were downloaded from the RCSB Protein Data Bank ~\cite{ref6}. PDB 1A3N is a 1.8 \AA{} X-ray structure of deoxy human hemoglobin; chains A and B were used for the $\alpha$ and $\beta$ subunits. PDB 3RGK is a 1.65 \AA{} X-ray structure of the human myoglobin K45R variant; its single sequence mismatch relative to the UniProt target was excluded from coordinate matching. AlphaFold Database models were retrieved through the public prediction API for UniProt P69905, P68871, and P02144 ~\cite{ref7}. The retrieval manifest records model URLs, database version, model creation date, global confidence, and SHA-256 hashes for every downloaded file.

PDB files were parsed from \nolinkurl{ATOM} records. We retained the first blank/A alternate-location C$\alpha$ atom for each residue in the declared chain. Three-letter residue names were mapped to one-letter codes. Predicted and experimental sequences were globally aligned with Needleman--Wunsch dynamic programming (match 2, mismatch $-$1, gap $-$2); coordinate pairs were retained only when both aligned residues were present and identical.

Predicted C$\alpha$ coordinates were superposed on experimental coordinates using the Kabsch least-squares rotation ~\cite{ref8}. We calculated whole-chain C$\alpha$ RMSD and a predeclared confidence-masked RMSD using matched AlphaFold residues with pLDDT$\geq$70. The rigid transform fitted to that high-confidence core was then applied to all matched residues before residue-level discrepancy screening. Moving-block residue bootstrap intervals used 2,000 resamples, blocks of 10 contiguous residues, and a fixed seed; because residues are not independent biological replicates, these intervals are descriptive. We also calculated median residue error, fractions within 2 and 5 \AA{}, mean pLDDT, and Spearman rank correlation between pLDDT and negative core-aligned residue error.

The expanded panel contains 15 human proteins and 2,688 matched residues. Experimental-method and resolution fields were parsed from PDB headers. A discrepancy region required pLDDT$\geq$90 and core-aligned C$\alpha$ error$\geq$2 \AA{}; adjacent qualifying residues were grouped. Every output is labeled \nolinkurl{novelty_status=not_established} and requires review of constructs, mutations, ligands, oligomeric state, alternative structures, and experimental conditions before independent validation. The Kabsch and core-transform implementations were regression-tested against synthetic rigid transformations. The analysis script, panel declaration, raw coordinate files, residue-level CSV, target summaries, source hashes, candidate table, and figures are included in the companion repository.

No correction for multiple comparisons was applied to the three correlation tests because the correlations are descriptive secondary endpoints in a small validation case study. Exact nominal P values are reported and no biological discovery claim is based on them. Local distance difference test (lDDT) is a widely used superposition-free structural score ~\cite{ref9}, but this compact study reports the simpler residue-level aligned C$\alpha$ error so every calculation remains inspectable in the supplied script.

\section{Results}

\subsection{Audit and repaired measurement paths}

The audit confirmed that the repository contains explicit idea/method and paper graphs, biology-specific routing, citation and evidence components, execution backends, manifests, and submission-package primitives. It also found that the default evaluation harness stops after idea and method generation; paper scoring therefore falls back to those texts when no paper exists. The autonomous loop adapters score existing artifacts but do not run a declared research pipeline. Paper reviewer roles currently use the drafting client. These findings constrain the interpretation of the present work: architecture and deterministic controls are evaluated; live end-to-end scientific performance and autonomous improvement are not.

After the three measurement repairs, targeted tests confirmed that biology task construction preserved the biology domain, method-signal recall was emitted, and real evidence-node JSONL artifacts contained Claim denominators and support links. The corrected artifact can now yield a nonzero unsupported-claim rate when drafted claims lack support, rather than a false 0.0 caused by an absent denominator.

\subsection{Software validation results}

The complete Python suite produced no failures or errors. The targeted biology-domain, genomics-adapter, evidence/citation, and adversarial-safety suites also produced no failures or errors. Some full-suite tests were skipped because optional live dependencies, provider credentials, or a platform-specific path condition were unavailable. Figure 2 and Table 1 report exact counts from the machine-readable validation report generated for this revision.

\begin{figure}[H]
\centering
\includegraphics[width=0.96\textwidth]{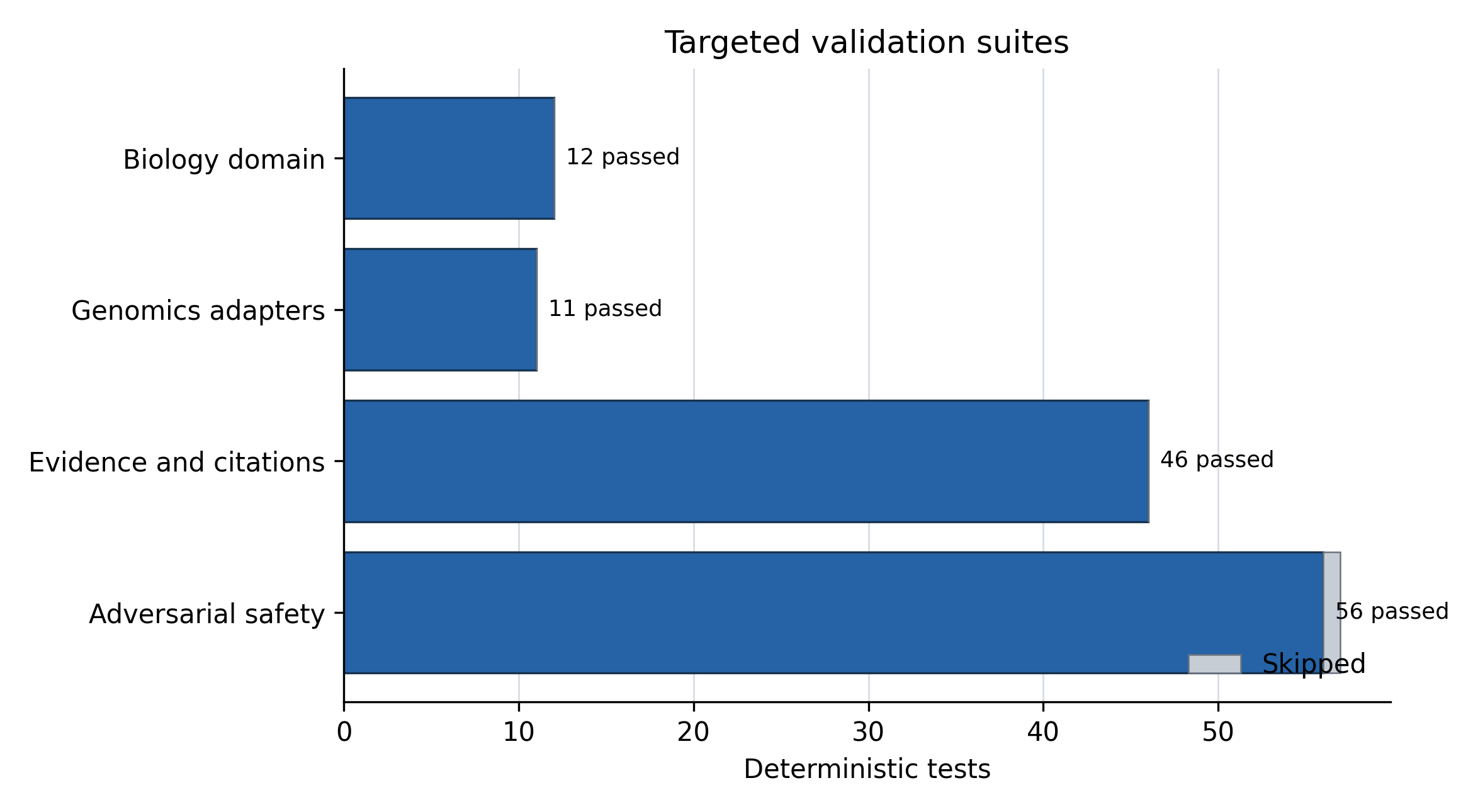}
\caption{Passed and skipped tests in targeted validation suites.}
\label{fig:figure-2}
\end{figure}

\begin{table}[H]
\centering
\caption{Deterministic software-validation results. Targeted suites overlap with the full Python suite.}
\label{tab:table-1}
\small
\resizebox{\textwidth}{!}{%
\begin{tabular}{lrrrr}
\toprule
Suite & Tests & Passed & Skipped & Failures/errors \\
\midrule
Biology domain & 12 & 12 & 0 & 0 \\
Genomics adapters & 11 & 11 & 0 & 0 \\
Evidence and citations & 46 & 46 & 0 & 0 \\
Adversarial safety & 57 & 56 & 1 & 0 \\
Full Python suite & 937 & 931 & 6 & 0 \\
\bottomrule
\end{tabular}%
}
\end{table}

The targeted suites are subsets of the full suite. These results support the narrower statement that the asserted software behaviors passed in the recorded environment. They do not measure paper correctness, novelty, human usefulness, or the probability that an open-ended agent run will succeed.

\subsection{Temporal rediscovery benchmark}

The synthetic engineering smoke behaved as intended. Across five tasks, frequency ranking yielded MRR 0.500 and Recall@1 0.000, while TF--IDF, bridge-only, and evidence-aware conditions each yielded MRR 1.000 and Recall@1 1.000. Recall@10 was 1.000 and the false-novelty rate on declared known controls was 0.000 under every condition. Because target evidence and decoys were constructed to test the implementation, these values are regression evidence, not biological efficacy.

The historical pilot produced a harder ranking (Table 2). Frequency ranked the held-out fish-oil/Raynaud relation third (reciprocal rank 0.333), TF--IDF ranked it second (0.500), and both bridge-only and evidence-aware conditions ranked it first (1.000). The candidate had one independent evidence path through blood viscosity, no direct pre-1986 prior record in the frozen corpus, complete URL/date provenance, and validation PMID 2536517 published in 1989. The three direct Raynaud--nifedipine records and one Raynaud--prostaglandin E1 record were correctly labeled known pre-cutoff, so none contributed a false-novelty error. The result supports the value of an explicit evidence bridge on this historical case; it does not estimate performance across biomedical discovery.

\begin{table}[H]
\centering
\caption{Historical temporal rediscovery pilot. The benchmark contains one manually curated task.}
\label{tab:table-2}
\small
\resizebox{\textwidth}{!}{%
\begin{tabular}{lrrrrr}
\toprule
Condition & Tasks & MRR & Recall@1 & Recall@10 & False-novelty rate \\
\midrule
Frequency & 1 & 0.333 & 0.000 & 1.000 & 0.000 \\
TF--IDF & 1 & 0.500 & 0.000 & 1.000 & 0.000 \\
A--B/B--C bridge & 1 & 1.000 & 1.000 & 1.000 & 0.000 \\
Evidence-aware & 1 & 1.000 & 1.000 & 1.000 & 0.000 \\
\bottomrule
\end{tabular}%
}
\end{table}

\begin{figure}[H]
\centering
\includegraphics[width=0.96\textwidth]{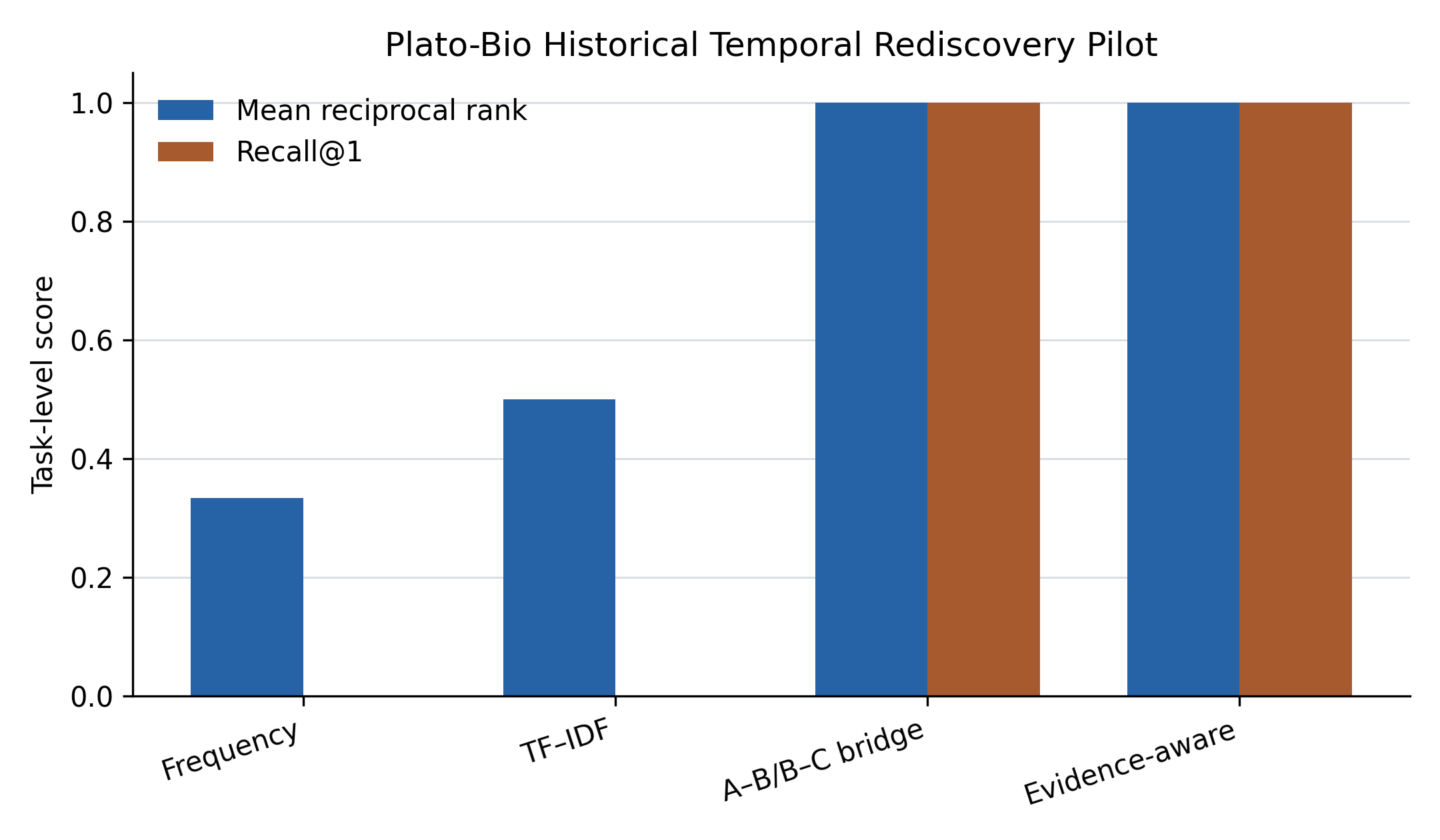}
\caption{Historical temporal rediscovery pilot across four ranking conditions.}
\label{fig:figure-3}
\end{figure}

\subsection{Original globin structural agreement}

All three targets produced high sequence coverage and sub-ångström C$\alpha$ RMSD after sequence-aware superposition (Table 3). Hemoglobin $\alpha$ had the lowest RMSD (0.270 \AA{}), followed by myoglobin (0.501 \AA{}) and hemoglobin $\beta$ (0.520 \AA{}). All matched $\alpha$ and $\beta$ hemoglobin residues were within 2 \AA{} of their experimental coordinates. For myoglobin, 146 of 147 matched residues (99.3\%) were within 2 \AA{} and all were within 5 \AA{}.

\begin{table}[H]
\centering
\caption{AlphaFold-to-experiment structural agreement in the declared globin panel.}
\label{tab:table-3}
\small
\resizebox{\textwidth}{!}{%
\begin{tabular}{lrrrrrrrr}
\toprule
Target & UniProt & PDB chain & Matched residues & Sequence identity & C$\alpha$ RMSD (\AA{}) & Median error (\AA{}) & Within 2 \AA{} & Mean pLDDT \\
\midrule
Hemoglobin $\alpha$ & P69905 & 1A3N A & 141 & 1.000 & 0.270 & 0.220 & 1.000 & 98.30 \\
Hemoglobin $\beta$ & P68871 & 1A3N B & 145 & 1.000 & 0.520 & 0.453 & 1.000 & 97.55 \\
Myoglobin K45R comparison & P02144 & 3RGK A & 147 & 0.987 & 0.501 & 0.310 & 0.993 & 97.71 \\
\bottomrule
\end{tabular}%
}
\end{table}

\begin{figure}[H]
\centering
\includegraphics[width=0.96\textwidth]{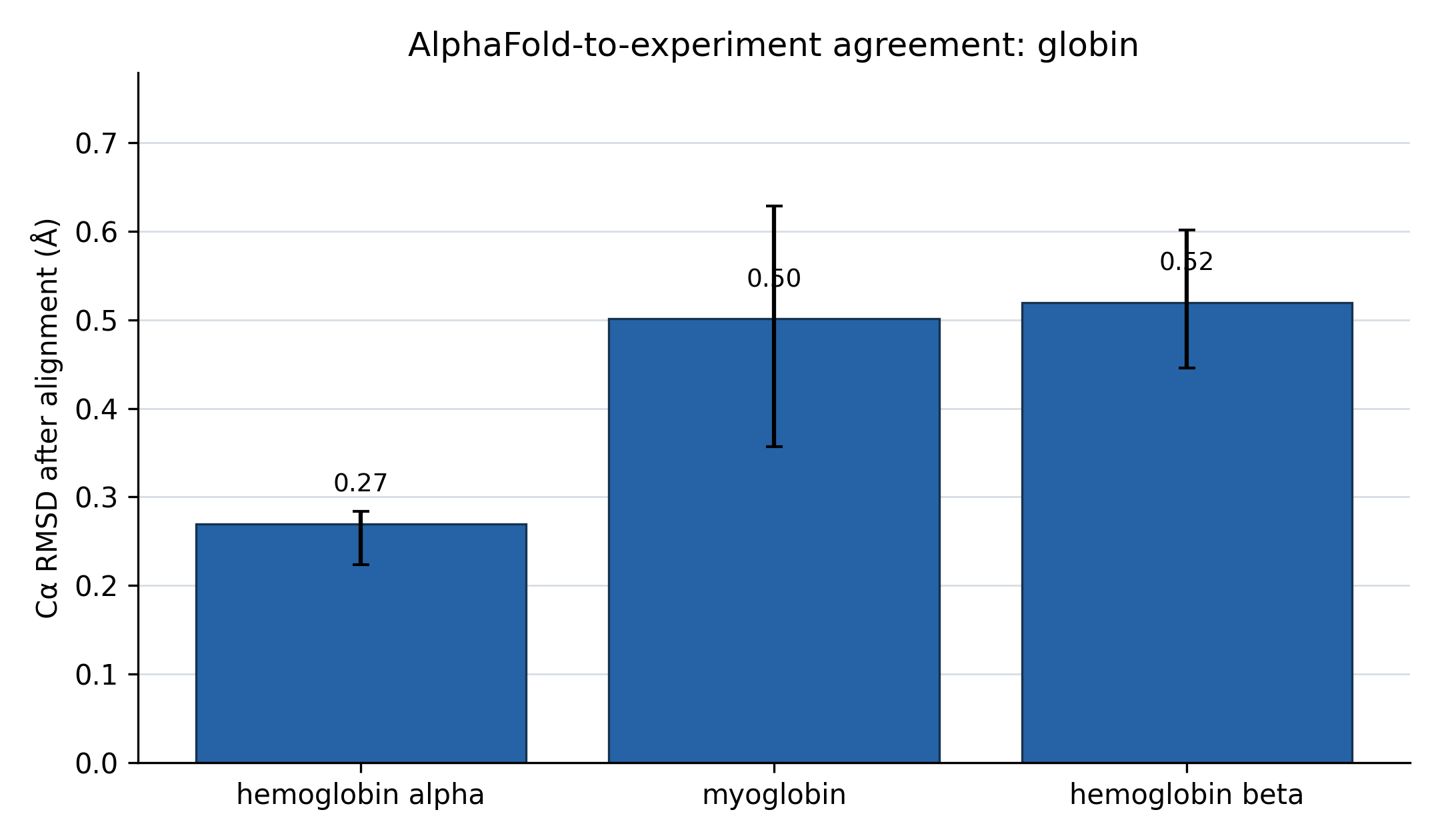}
\caption{C$\alpha$ RMSD for the three declared globin targets.}
\label{fig:figure-4}
\end{figure}

Residue-level pLDDT was positively associated with lower aligned coordinate error in each target: hemoglobin $\alpha$ $\rho$=0.242 (P=0.00388), hemoglobin $\beta$ $\rho$=0.353 (P=1.34$\times$$10^{-5}$), and myoglobin $\rho$=0.283 (P=0.000515). The restricted pLDDT range (mean 97.5--98.3) limits calibration inference, and the correlations should be interpreted as descriptive consistency rather than proof that pLDDT is fully calibrated for this family.

\begin{figure}[H]
\centering
\includegraphics[width=0.96\textwidth]{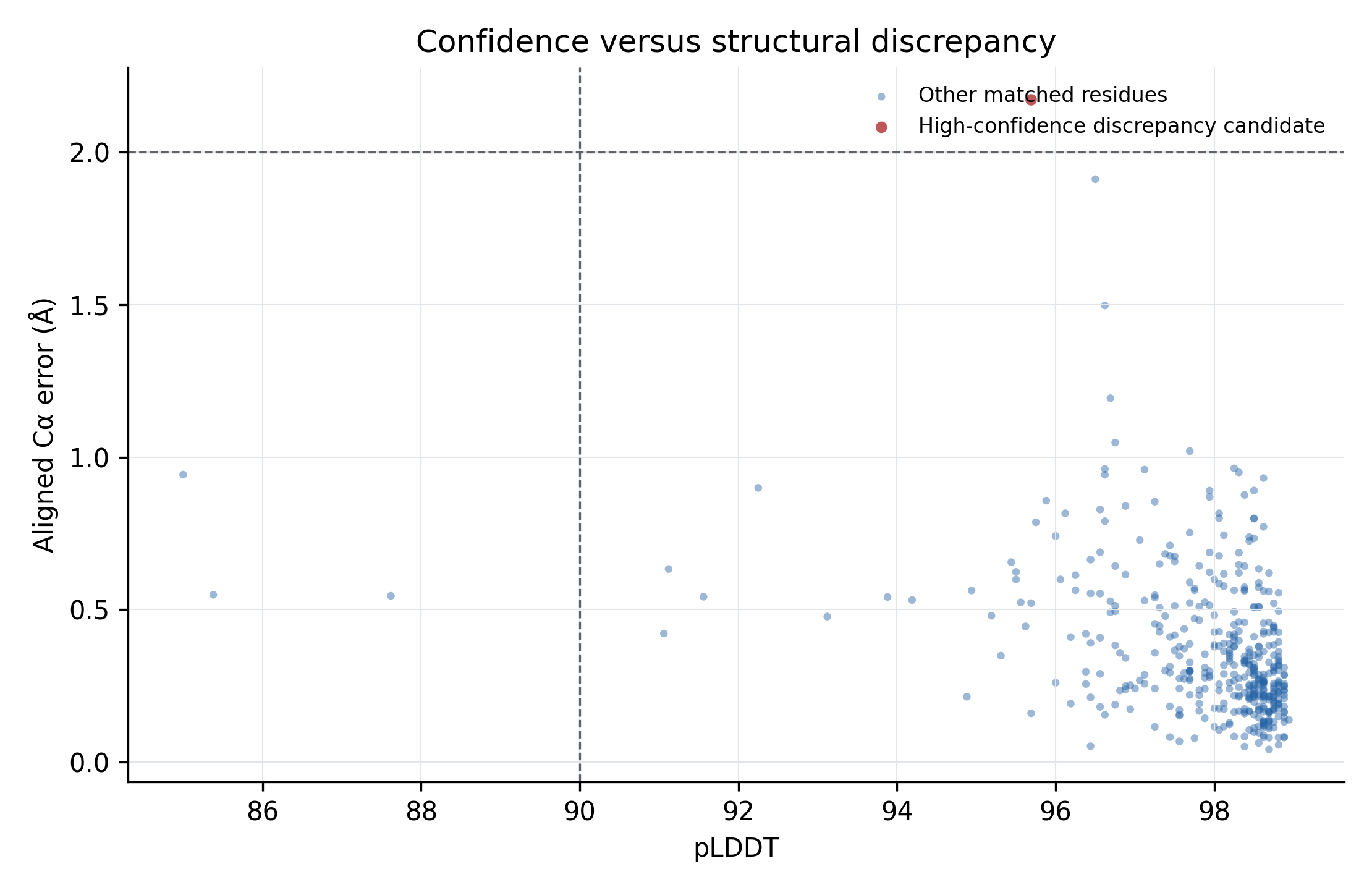}
\caption{Residue-level pLDDT versus aligned C$\alpha$ error.}
\label{fig:figure-5}
\end{figure}

\subsection{Diverse structural screen and hypothesis triage}

All 15 declared targets completed without retrieval or analysis failure. The panel contained 2,688 sequence-matched residues; median whole-chain RMSD was 0.520 \AA{} and 9 of 15 targets were below 1 \AA{}. After fitting only residues with pLDDT$\geq$70, median core RMSD was 0.501 \AA{} and 11 of 15 targets were below 1 \AA{} (Figure 6). Compact proteins such as transthyretin, hemoglobin $\alpha$, cytochrome c, superoxide dismutase 1, cyclophilin A, carbonic anhydrase 2, and lysozyme ranged from 0.258 to 0.420 \AA{} in the high-confidence-core comparison. CDK2 and MAPK1 improved from whole-chain RMSDs of 1.873 and 1.705 \AA{} to core values of 0.786 and 0.740 \AA{}, respectively, indicating that lower-confidence regions dominated much of their global discrepancy.

Four targets retained high-confidence-core RMSD above 2 \AA{}: KRAS (2.085 \AA{}), SUMO1 (2.576 \AA{}), TP53 (3.889 \AA{}), and estrogen receptor $\alpha$ (4.961 \AA{}). These comparisons differ in construct coverage, domain state, ligands, and experimental modality. SUMO1 is a solution-NMR ensemble with a flexible N-terminus; fitting the 74 high-confidence residues reduced its RMSD from 16.610 \AA{} over the whole chain to 2.576 \AA{}. TP53 and estrogen receptor $\alpha$ experimental chains cover only 49.4\% and 40.0\% of their respective full AlphaFold sequences. These facts make the high values informative triage signals but inadequate evidence of a novel conformation.

The predeclared rule emitted 27 high-confidence discrepancy regions, 9 containing at least two adjacent residues. The candidates are machine-readable and traceable to residue coordinates, but none is called a discovery. The most defensible immediate finding is methodological: confidence masking and experimental-context stratification prevent flexible termini and partial-domain constructs from dominating novelty screens, while retaining a short list for targeted structural review.

\begin{figure}[H]
\centering
\includegraphics[width=0.96\textwidth]{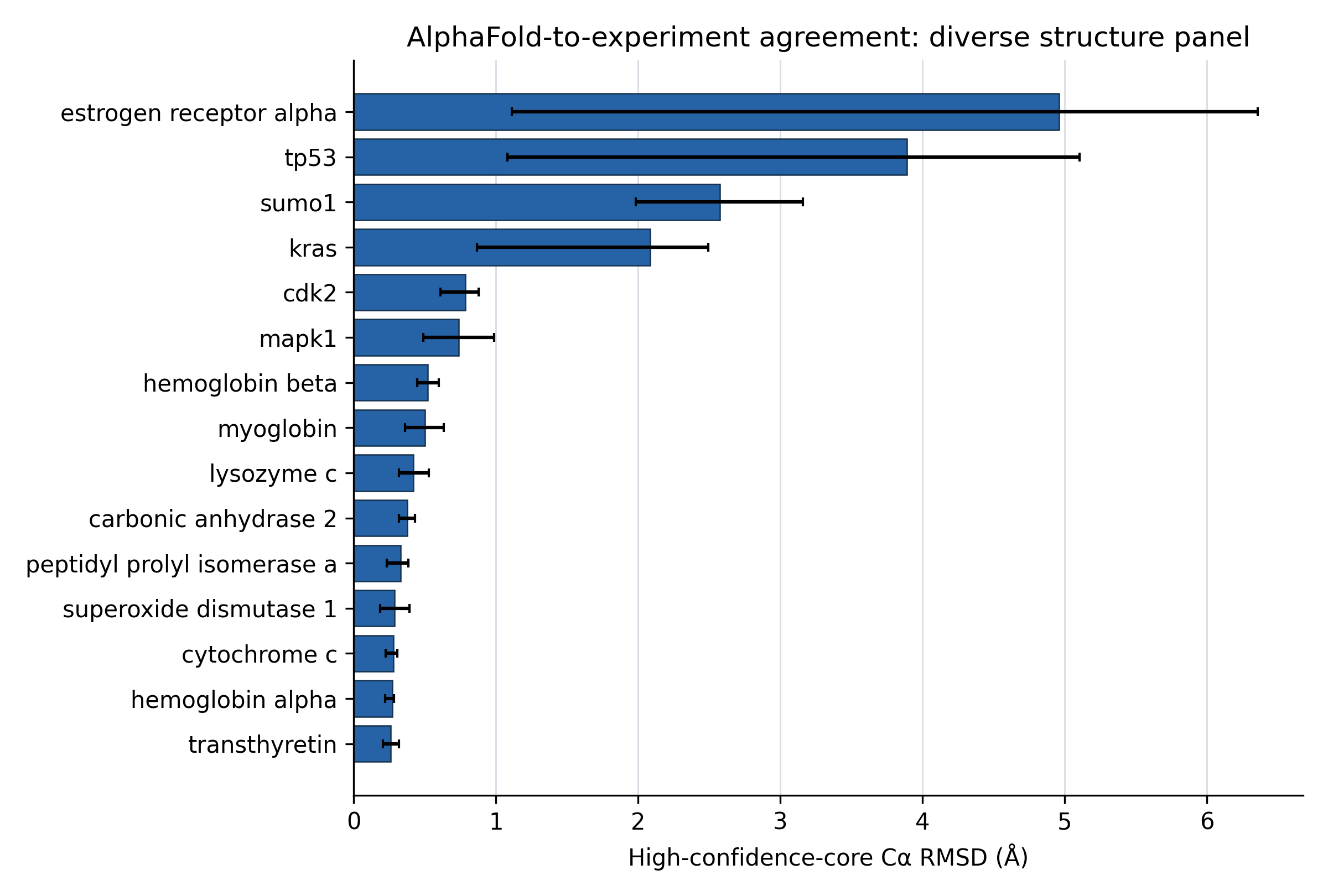}
\caption{High-confidence-core C$\alpha$ RMSD across the declared 15-target structural panel.}
\label{fig:figure-6}
\end{figure}

\subsection{Reproducibility artifacts}

The evidence bundle contains the original globin inputs, 15-target panel declaration, 29 cached coordinate files for the diverse screen (15 AlphaFold models and 14 unique experimental PDB entries), source and output hashes, target- and residue-level CSV files, 27 discrepancy candidates, frozen temporal fixtures, complete candidate rankings, bootstrap summaries, and JSON manifests. The pinned CompBioBench catalog is stored separately from Plato-Bio results so imported task metadata cannot be mistaken for executed performance. Re-running the supplied scripts reconstructs the tables and Figures 3--6. The paper source, editable Word manuscript, LaTeX package, and submission PDFs are generated from the same controlled bundle.

\section{Discussion}

The audit and benchmarks point to a practical lesson: visible workflow stages do not by themselves make an agent's output correct. Plato-Bio instead separates source records, citation reports, claims, evidence links, analysis outputs, manifests, and manuscript files so that each can be inspected on its own. The repaired mixed JSONL contract shows why this separation matters. Without persisted claims, an unsupported-claim metric could report a reassuring zero even though the denominator required to calculate it was absent.

The biology domain profile and genomics registry provide an extensible interface for biological work. Domain routing reduces accidental use of astronomy-specific sources, while typed adapters expose requirements, permissions, coordinate conventions, and expected artifacts before an external genomics command is executed. This is useful safety and reproducibility scaffolding, but it is not a validation of GenomeKit, Paragraph, ExpansionHunter Denovo, Gauchian, ZIPPY, or Illumina Connected Analytics themselves.

The globin study demonstrates the intended evidence pattern. Source identities are fixed, coordinates are archived with hashes, alignment rules are explicit, the rigid-body fit is regression-tested, target- and residue-level results are preserved, and conclusions are proportional to the data. The sub-ångström agreement is consistent with the strong performance reported for AlphaFold and with the purpose of AlphaFold DB ~\cite{ref7,ref10}. It should not be interpreted as a new benchmark of the broader AlphaFold proteome: the panel contains three related, high-confidence globins selected for a compact reproducibility case study, and one experimental structure contains a known point mutation.

The temporal pilot addresses a different question: whether a candidate relation can be reconstructed from literature that predates its direct evaluation. On the Raynaud task, generic frequency and lexical relevance favored known or more frequently mentioned concepts, whereas the independent evidence bridge recovered the held-out relation. The result is mechanistically interpretable because every score can be traced to PMID 58309 and PMID 4015748 and because direct pre-cutoff treatment records remain visible as known controls. This is a stronger novelty-screening contract than treating linguistic dissimilarity as novelty. It remains a retrospective case constructed with knowledge of the later literature, so prospective candidates must be reviewed against broader corpora and independently tested.

The 15-target screen shows why structural novelty triage needs confidence and context. Whole-chain RMSD can be dominated by disordered segments, while partial constructs and ligand-dependent states can produce legitimate high-confidence differences that are already known. Reporting both whole-chain and high-confidence-core values preserved these distinctions. The 27 emitted regions are therefore useful as an auditable review queue, not as 27 findings. A true structural novelty claim would require alternate-structure searches, construct and ligand matching, domain-aware or local scores, ensemble analysis, and ideally new experimental evidence.

The source audit also changes how the agent itself should be described. The current default evaluation is an idea/method benchmark, not an end-to-end paper benchmark. The paper's reviewer roles are same-model self-critique, not independent review. The autonomous loop infrastructure can keep or discard scored states, but its default adapters do not execute a full research cycle. These limitations are not wording details; they determine what can be concluded from future evaluations. A definitive efficacy study will require frozen biological tasks, versioned datasets, real results and paper generation, multiple stochastic repetitions, baseline/ablation conditions, and blinded domain-expert adjudication.

\subsection{Limitations}

The temporal historical benchmark contains one manually curated task. Its bootstrap interval is necessarily degenerate, candidate concepts are not an exhaustive search space, concept annotations were curated with knowledge of the historical relation, and PubMed records outside the frozen set could change the rank or prior-art label. The five synthetic tasks are implementation fixtures. Generalization requires a preregistered set of many historical tasks curated by multiple experts with adjudicated cutoffs, broader frozen retrieval, harder decoys, and blinded evaluation. No prospective candidate was generated or experimentally validated.

The structural screen remains small and does not include cryo-electron microscopy density fitting, lDDT, TM-score, side-chain accuracy, ligand geometry, oligomeric interfaces, domain segmentation, or conformational-ensemble comparison. The high-confidence mask is based on AlphaFold pLDDT rather than an independently selected structural domain. PDB 3RGK is a K45R myoglobin mutant; 1A5R is an NMR ensemble represented by its first parsed model; and several PDB chains are partial constructs. Correlation tests and residue-bootstrap intervals are descriptive. Raw structures were retrieved from public services whose upstream records may be revised; file hashes freeze the exact analyzed bytes.

The software evaluation is deterministic and largely unit-level. LLM calls, paid retrieval services, Modal, E2B, hosted PostgreSQL, and authenticated Hugging Face paths were not available in the recorded environment. No claim is made that the configured citation threshold equals observed citation accuracy. Prompt-injection screening and scientific-verifier rules are heuristic and do not sandbox generated code or prove factual correctness. Local execution of LLM-generated code should not be used with sensitive data or production credentials without an external sandbox and human review.

Authorship and submission metadata remain human responsibilities. The corresponding author must confirm the author list, affiliations, ORCIDs, contribution statement, funding, conflicts, and consent before deposit. arXiv posting and moderation do not constitute peer review, and this preprint should not be represented as certified scientific evidence.

\section{Conclusion}

Plato-Bio provides an inspectable foundation for supervised computational-biology workflows and a more defensible route from retrieved literature or predicted structures to candidate hypotheses. Independent A--B/B--C evidence bridges outperformed frequency and TF--IDF in one historical rediscovery pilot, while confidence-masked structural comparison separated compact sub-ångström agreement from context-sensitive discrepancies across 15 proteins. The system now emits traceable candidates, known controls, abstentions, and explicit \nolinkurl{not_established} novelty labels instead of treating difference as discovery. The evidence supports software-contract reproducibility, one retrospective literature case, and descriptive structural triage. Establishing general agent efficacy or biological novelty still requires a larger preregistered benchmark, independent expert assessment, prospective analysis, and experimental validation.

\section{Data and code availability}

Code, manuscript source, experiment scripts, exact input coordinate files, hashes, derived CSV/JSONL files, figures, and validation manifests are available in the companion repository: \url{https://github.com/Eldergenix/Plato-Scientific-Research-Autonomous-Agent}. The software validation reported here was run from the clean repository state at commit \nolinkurl{d809f2b}; the full SHA is recorded in the validation manifest. The original globin bundle is under \nolinkurl{preprint/results/globin_benchmark/}; the expanded structural screen is under \nolinkurl{preprint/results/diverse_structure_benchmark/}; and temporal fixtures and results are under \nolinkurl{evals/biological_novelty/fixtures/} and \nolinkurl{preprint/results/temporal_novelty_*}. The pinned, hash-verified CompBioBench catalog is metadata only and is stored under \nolinkurl{preprint/results/compbiobench_catalog/}.

\section{Ethics statement}

This study used public protein-structure records and software tests. It involved no human participants, identifiable private information, animals, or prospective clinical intervention; institutional ethics review was not required. The software must not be used to make clinical decisions without independent validation and appropriate regulatory and ethical oversight.

\section{Funding}

No external funding was reported for this study.

\section{Competing interests}

The author maintains the Eldergenix fork of Plato described in this manuscript. No other competing interests were reported.

\section{Author contributions}

S.C.: conceptualization, software, methodology, investigation, validation, data curation, visualization, writing---original draft, writing---review and editing, and project administration. Original Denario/Plato contributors are credited through citation ~\cite{ref4} and the software repository history; they are not listed as authors of this fork-specific study without explicit authorship consent.

\section{Generative AI disclosure}

OpenAI Codex (GPT-5) assisted with repository auditing, experimental scripting, test repair, figure preparation, and manuscript drafting. The human author is responsible for the study design, verification of analyses, accuracy, originality, interpretation, and final submitted text. The AI system is not an author.

\section{Acknowledgments}

We thank the original Denario/Plato contributors for releasing the foundational research-agent code, the RCSB Protein Data Bank for experimental structures, and EMBL-EBI and Google DeepMind for AlphaFold DB predictions.

\section{Figure legends}

\textbf{Figure 1. Plato-Bio architecture and verification boundaries.} The main research stages are represented as an explicit graph. Lower controls emit or validate artifacts used by downstream stages. The diagram describes implemented topology and does not imply that each stage has been empirically validated end to end.

\textbf{Figure 2. Targeted deterministic validation suites.} Bars show passing tests in control-specific subsets; gray denotes skipped tests. Suites overlap with the full repository suite and should not be summed.

\textbf{Figure 3. Historical temporal rediscovery pilot.} Bars compare mean reciprocal rank and Recall@1 for frequency, TF--IDF, A--B/B--C bridge, and evidence-aware ranking. The benchmark contains one manually curated task, so the values are descriptive and do not estimate general discovery performance.

\textbf{Figure 4. AlphaFold-to-experiment agreement in the declared globin panel.} Bars show global C$\alpha$ RMSD after sequence-aware matching and Kabsch superposition. Values are calculated from the supplied target summary.

\textbf{Figure 5. Residue confidence versus aligned coordinate error.} Each point is one matched residue. pLDDT is read from the AlphaFold PDB B-factor field; error is the Euclidean C$\alpha$ distance after target-level superposition.

\textbf{Figure 6. Diverse structural screen.} Bars show C$\alpha$ RMSD after Kabsch fitting on matched AlphaFold residues with pLDDT$\geq$70. Error bars are descriptive 95\% moving-block residue-bootstrap intervals and are not population-level confidence intervals.

\end{document}